\documentclass[runningheads]{llncs}

\usepackage{amsmath}
\usepackage{amssymb}
\usepackage{booktabs}
\usepackage{array}
\usepackage{multirow}
\usepackage{rotating}
\usepackage[caption=false]{subfig}

\begin{document}

\title{When Collaborative Filtering Meets Reinforcement Learning}

%%
%\title{Contribution Title\thanks{Supported by organization x.}}
%%


\author{Yu Lei \and Wenjie Li}


\institute{The Hong Kong Polytechnic University \\
			 Hong Kong, China\\
	\email{\{csylei, cswjli\}@comp.polyu.edu.hk}}

%\institute{Princeton University, Princeton NJ 08544, USA \and
%Springer Heidelberg, Tiergartenstr. 17, 69121 Heidelberg, Germany
%%ABC Institute, Rupert-Karls-University Heidelberg, Heidelberg, Germany\\

\maketitle              % typeset the header of the contribution

\begin{abstract}
In this paper, we study a multi-step interactive recommendation problem, where the item recommended at current step may affect the quality of future recommendations. To address the problem, we develop a novel and effective approach, named CFRL, which seamlessly integrates the ideas of both collaborative filtering (CF) and reinforcement learning (RL). More specifically, we first model the recommender-user interactive recommendation problem as an agent-environment RL task, which is mathematically described by a Markov decision process (MDP). Further, to achieve collaborative recommendations for the entire user community, we propose a novel CF-based MDP by encoding the states of all users into a shared latent vector space. Finally, we propose an effective Q-network learning method to learn the agent's optimal policy based on the CF-based MDP. The capability of CFRL is demonstrated by  comparing its performance against a variety of existing methods on real-world datasets.

\end{abstract}

\section{Introduction}
Collaborative filtering (CF) is one of prominent techniques to build personalized recommender systems, which has been successfully applied by many web applications \cite{ricci2011introduction}. In particular, the model-based CF approaches that utilize latent factor models, have shown powerful capability for user preference modeling, and demonstrated high accuracy and scalability for recommendations in large datasets \cite{Koren2009,he2016fast}. Despite the success of existing approaches, they mainly focus on providing accurate recommendations for only one round of recommender-user interaction. However, in more practical scenarios, the recommender usually interacts with the user for multiple rounds, and the recommendations provided in current round may affect the quality of future recommendations. For example, the items recommended in earlier rounds may not be liked by the user, but the received feedbacks provide useful information which can help the recommender make better recommendations in later rounds. For such scenarios, the existing approaches ignore the influences from earlier rounds to later ones, which can only make locally optimal recommendations for each round. Instead, a more desired approach should focus on providing globally optimal recommendations for all possible rounds. 

Aiming at this issue, a potential solution is to utilize the techniques of reinforcement learning (RL) \cite{sutton1998reinforcement} to model the multi-round recommendations as a multi-step decision making problem. RL has been proved that has the ability to make optimal multi-step decisions for many complex problems such as playing Atari \cite{mnih2015human} and the game of Go \cite{silver2016mastering}. By using RL, an intelligent agent can be learned to recommend an appropriate item at each step, so as to maximize the globally optimal recommendation performance for all steps. In the literature, RL-based approaches have been proposed to solve a number of recommendation problems such as session-based recommendation \cite{shani2005mdp,zhao2018deep,zhao2018recommendations}, news article recommendation \cite{zheng2018drn}, and music playlist recommendation \cite{hu2017playlist}. However, these approaches are only applicable to implicit-feedback recommender systems. Besides, they fail to model neighbors' collective preferences of the target user, which cannot provide collaborative recommendations for the entire user community.

Distinct from the existing work, we study an interactive recommendation problem for explicit-feedback recommender systems, and seek to develop an RL-based collaborative recommendation approach. We first model the recommender-user interactive recommendation problem as an agent-environment RL task, which is mathematically described by a Markov decision process (MDP). The MDP formulation makes it reasonable and feasible to employ standard RL methods such as Q-learning to learn the agent's optimal policy. Further, to achieve collaborative recommendations, we propose a CF-based MDP by encoding the states of all users into a shared latent vector space. More specifically, by utilizing the technique of matrix factorization \cite{Koren2009}, the state of a user at a given time step is represented by a low-dimensional feature vector that describes the observed user preference. Such CF-based states enable the agent to learn a collaborative recommendation policy based on all users' data. Finally, to solve the CF-based MDP, we propose a Q-network learning method based on deep Q-learning (DQN) \cite{mnih2015human}, with a particular training scheme that uniformly samples state transitions from different users. We name the proposed approach CFRL, since it integrates the ideas of both CF and RL. We empirically validate the performance of CFRL on three real-world datasets, compared to a wide variety of existing approaches, including active learning (AL), collaborative filtering (CF), multi-armed bandit (MAB) and reinforcement learning (RL) methods. The experimental results demonstrate the capability of CFRL for interactive recommendations. In particular, due to the use of CF-based states, CFRL shows overwhelming advantage over DQN.

\section{Problem Definition}
Suppose we have a 5-star recommender system with integer ratings in $\{1,2,3,4,5\}$, which currently involves $m$ users and $n$ items. Let $\mathcal{U}=\{1,...,m\}$ and $\mathcal{I}=\{1,...,n\}$ denote the sets of users and items, respectively. Let $R\in\mathbb{R}^{m\times n}$ denotes the observed user-item rating matrix, where each nonzero $R_{ui}$ denotes the observed rating of item $i$ given by user $u$, and each zero implies that the user has not rated the item yet. We consider a recommendation scenario of cold-start user as follows. Suppose a new user $u=m+1$ enters into the system at time step $t=0$. The recommender provides an item to the user, then receives a rating on the item given by the user. After considering the observed rating, the recommender updates its knowledge about the user and provides a new item at time step $t=1$. Suppose such a recommender-user interactive process lasts for $T$ time steps. The goal of the recommender is to recommend the most interesting items that can maximize the sum (or average) of ratings received over $T$ steps.

\section{The Proposed Approach: CFRL}
In this section, we present our approach, CFRL, for $T$-step interactive recommendation. We first formulate the problem as an MDP. Then, we describe a CF-based state representation method to encode the states into latent space. Finally, we propose a Q-network learning method to solve the CF-based MDP.

\subsection{Formulating the Problem as an MDP}

We consider the aforementioned interactive recommendation problem under the standard RL framework. The recommender-user interaction in recommendation can be naturally modeled as the agent-environment interaction in RL. At each time step $t$, the agent (recommender) observes a state $s_t$ about the environment (user $u$), then takes an action (item) $a_t$ according to its policy $\pi$, which is usually a mapping from states to action probabilities. One time step later, as a result of its action, the agent receives a numerical reward (rating) $r_{t+1}$ and a new state $s_{t+1}$ from the environment. The goal of the agent is to maximize the cumulative reward it receives over $T$ time steps. According to \cite{sutton1998reinforcement}, such an RL task can be mathematically described by an MDP, a tuple $(\mathcal{S}, \mathcal{A}, \mathcal{P}, \mathcal{R})$ defined as follows.

\textbf{$\mathcal{S}$ is the state space.} The state $s_t$ represents the observed preference of user $u$ at time step $t$. A straightforward state representation method is to define the state $s_t$ as a $n$-dimensional rating vector $R_{u*}^{(t)}$, which denotes the $u$-th row of $R$ at time step $t$.  The nonzero values of $R_{u*}^{(t)}$ indicate the observed ratings given by user $u$. Obviously, the initial state $s_0$ is a zero vector. 

\textbf{$\mathcal{A}$ is the action space.} We define $\mathcal{A}$ as the set of all items, i.e., $\mathcal{A}=\mathcal{I}$. In each state $s_t$, an action $a_t$ can be taken from the set of available actions $\mathcal{A}(s_t)$, which is defined recursively: $\mathcal{A}(s_t)=\mathcal{A}(s_{t-1})\setminus\{a_{t-1}\}$ for $t\neq0$, and $\mathcal{A}(s_0)=\mathcal{A}$. In other words, the agent is not allowed to choose the items that have been recommended at previous time steps.

\textbf{$\mathcal{P}$ is the transition function.} $\mathcal{P}_{ss'}^a=Pr[s_{t+1}=s'|s_t=s,a_t=a]$ denotes the probability that the environment transits to state $s'$ after receiving action $a$ in state $s$. In the recommendation setting, the exact transition probabilities are unknown in advance. The agent can observe specific state transitions by interacting with the environment step by step. 

\textbf{$\mathcal{R}$ is the reward function.} $\mathcal{R}_{ss'}^a=E[r_{t+1}|s_t=s,a_t=a,s_{t+1}=s']$ denotes the expected immediate reward the environment generates after the transition from state $s$ to $s'$ due to action $a$. In the recommendation setting, the immediate reward of executing an action $a$ only depends on the rating given by user $u$. Therefore, we define $\mathcal{R}_{ss'}^a=R_{ua}$.

\subsection{Encoding the States into Latent Space}
With the MDP formulation, we can naturally employ standard RL methods such as Q-learning to learn the agent's optimal policy. However, the ultimate performance of the learned agent also depends on the quality of the state representations of the MDP. As pointed out in \cite{sutton1998reinforcement}, the state representations should capture sufficient and representative information about the states for specific tasks, in the sense that prior domain knowledge can be well embedded. For interactive recommendation, we believe that the state representations should effectively capture both the preference of a target user and the relationships between different users, so as to achieve personalized and collaborative recommendations for all users. Obviously, the raw state of the MDP, $s_t^o$ (we will use this notation in the rest of the paper), cannot meet the two requirements since it is high-dimensional and extremely sparse in practice.

Inspired by the idea of CF, we employ latent factor models to encode the sparse raw states to low-dimensional and dense feature representations. The latent factor models have been proved that have powerful capabilities on modeling the user preferences, the relationships between users, as well as the interdependencies between items \cite{Koren2009}. By mapping all users and items into the shared low-dimensional vector space, some effective latent features can be learned to model users' preferences and items' properties. Thus, it is natural and reasonable to utilize the latent feature vector of user $u$ to represent $u$'s states in the MDP. 

More specifically, we first pre-train a matrix factorization (MF) model based on the observed rating matrix $R\in\mathbb{R}^{m\times n}$ of training users, by minimizing a squared loss function \cite{Koren2009} defined as:
\begin{align}
	\label{eq_loss_mf}
	\mathcal{L} = \sum\nolimits_{u,i}(U_u^\mathrm{T}V_i - R_{ui})^2 + \lambda(\|U\|_\mathrm{F}^2 + \|V\|_\mathrm{F}^2),
\end{align}
where $\|\cdot\|_\mathrm{F}$ denotes the Frobenius norm, $\lambda$ is the regularization parameter, and $U\in\mathbb{R}^{d\times m}$ and $V\in\mathbb{R}^{d\times n}$ denote the latent feature matrices of users and items, respectively. 

The pre-trained item feature vectors $V_i\in\mathbb{R}^{d}$ for all items $i=1,...,n$, will be fixed and used to update the user feature vector $U_u\in\mathbb{R}^{d}$ for target user $u$. During the agent-environment interactive process, we continuously maintain the user feature vector $U_u$ over time steps, and use it as the new CF-based state $s_t$. Moreover, to ensure efficient online learning, $U_u$ is updated based on the latest observed rating $R_{ui}$ (which is located in the raw state $s_t^o$), by performing stochastic gradient descent (SGD) over the loss $\mathcal{L}$ in Equation \ref{eq_loss_mf}. The detailed algorithm of the proposed CF-based state representation method is presented in Algorithm 1. Similar to the online updating in \cite{he2016fast}, for the while loop in Algorithm 1, we find that one iteration is usually sufficient to achieve good results.

\begin{table}[t]
	\renewcommand\tabcolsep{1pt}
	\renewcommand\arraystretch{1.0}
	\centering
	\begin{tabular*}{\columnwidth}{rl} 
		\toprule
		\multicolumn{2}{l}{\textbf{Algorithm 1:} CF-based State Representation}\\
		\hline
		\multicolumn{2}{l}{\textbf{Input:} pre-trained $V$, vector $U_u$, learning rate $\alpha$, regularization $\lambda$, step $t$, raw state $s_t^o$}\\
		\multicolumn{2}{l}{\textbf{Output:} CF-based state $s_t$, vector $U_u$}\\
		1.& \textbf{if} $t=0$ \textbf{then} \\
		2.& \quad Initialize $U_u$ with zeros \\
		3.& \textbf{else} \\
		4.& \quad // Minimize $\mathcal{L}$ over the latest observed rating $R_{ui}$ \\
		5.& \quad \textbf{while} $\mathcal{L}$ is not converged \textbf{do} \\
		6.& \quad\quad $U_u\leftarrow U_u-2\alpha\left[(U_u^\mathrm{T}V_i - R_{ui})V_i+\lambda U_u\right]$ \\
		7.& \quad\quad $V_i\leftarrow V_i-2\alpha\left[(U_u^\mathrm{T}V_i - R_{ui})U_u+\lambda V_i\right]$ \\
		8.& \quad \textbf{end while}\\
		9.& \textbf{end if}\\
		10.& $s_t\leftarrow U_u$ \\
		\hline
	\end{tabular*}
\end{table}

With the CF-based state representation method, we actually derive a CF-based MDP with a low-dimensional continuous state space. Note that the CF-based MDP still has the Markov property, as the state $s_t$ is updated based on only the state $s_{t-1}$, which is independent of $s_{t-2},s_{t-3},...,s_0$. This indicates that we can employ standard RL methods to solve the CF-based MDP. Besides, other more complex CF-based models such as those proposed in \cite{Koren2009} can be easily incorporated into the framework to construct the CF-based states. The performance is supposed to be consistently improved as long as the adopted models can better capture the user preferences and the relationships between users. However, in this paper, we only use the standard MF model as an instantiation due to its simplicity, and focus on validating the capability of the general framework of our proposed approach.

\subsection{Q-network Learning for the CF-based MDP} 
To handle the CF-based MDP with continuous state space, we propose a Q-network learning method based on DQN \cite{mnih2015human}, which is essentially Q-learning with function approximation. We employ a feedforward neural network with weights $\textbf{w}$, referred to as Q-network $\hat{Q}(s,a,\textbf{w})$, as the function approximator to estimate the true action-value function $Q(s,a)$. The Q-network $\hat{Q}$ uses the CF-based state $s_t$ as input, and outputs the Q values of all possible actions in that state. To update the weights $\textbf{w}$ of $\hat{Q}$, stochastic gradient descent (SGD) can be used to minimize the mean squared error (MSE) between the predicted $\hat{Q}$ value and the Q-learning target $y$, for a given transition $(s_t, a_t, r_{t+1}, s_{t+1})$:
\begin{align}
	\textbf{w}\leftarrow \textbf{w} 
	+\alpha\left[y-\hat{Q}(s_t,a_t,\textbf{w})\right] \nabla_\textbf{w}\hat{Q}(s_t,a_t,\textbf{w}),
	\label{eq_qlearning_fa}
\end{align} 
where $\alpha$ is learning rate, and $y=r_{t+1}+\gamma\max_{a}\bar{Q}(s_{t+1},a)$ if $s_{t+1}$ is a non-terminal state and $y=r_{t+1}$ otherwise, and $\gamma$ is the discount factor that balances the importance between future rewards and immediate rewards. $\bar{Q}$ is called target network which copies the weights of $\hat{Q}$ regularly after $L$ steps during the interactive process. By continuing updates on all possible transitions, the learned $\hat{Q}$ will converge to the optimal action-value function $Q^{\ast}$, according to Bellman optimality equation \cite{sutton1998reinforcement}. The greedy policy with respect to $Q^{\ast}$ will be an optimal policy $\pi^{\ast}$.

\begin{figure}[t]
	\centering
	\includegraphics[width=0.7\columnwidth]{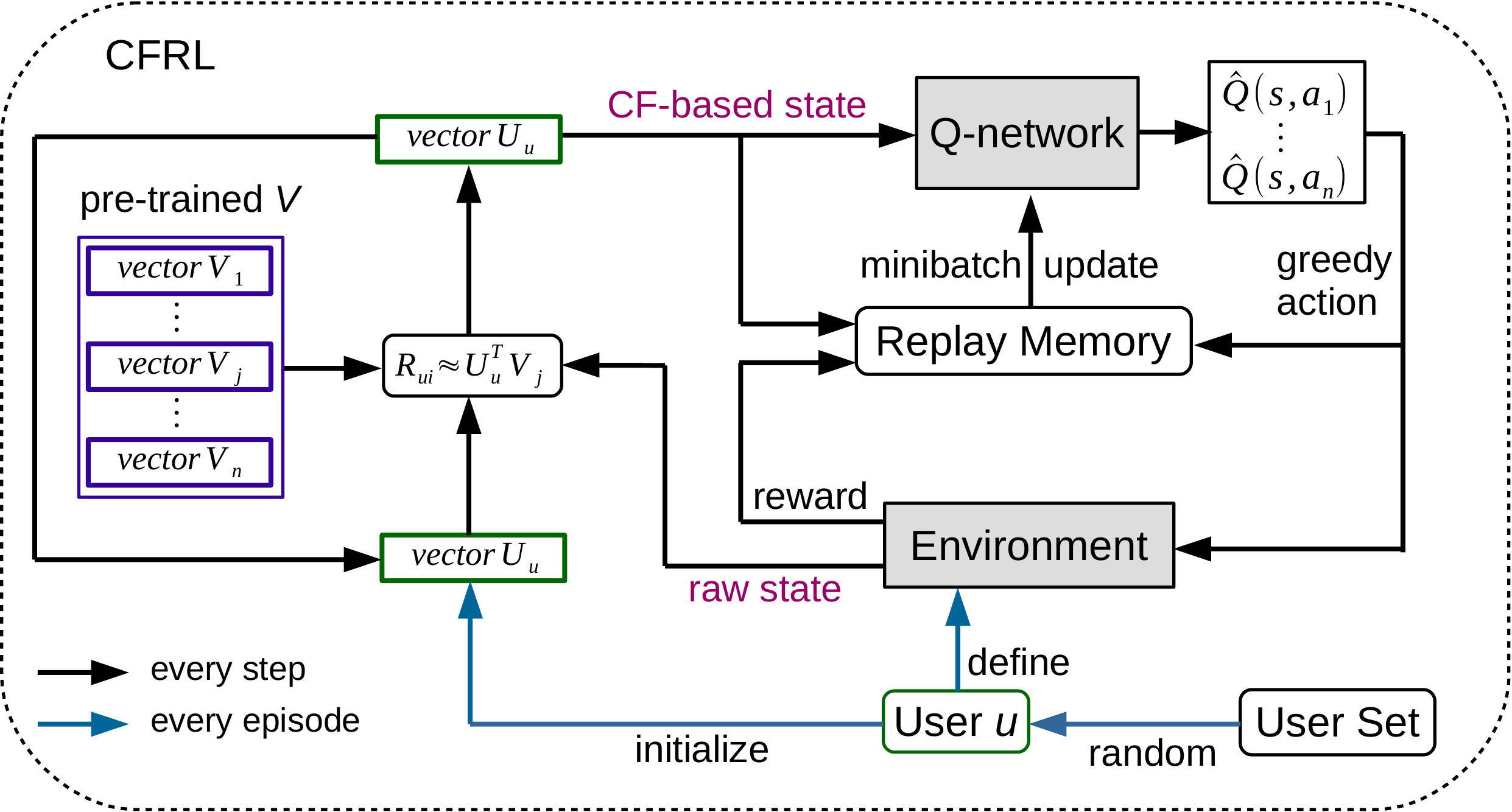}
	\caption{The basic framework of CFRL.}
	\label{fig_cfdqn}
\end{figure}

To sample sufficient transitions for Q-network learning, we propose a particular training scheme to train the CFRL agent based on the data of all training users. The basic framework and the learning algorithm of CFRL are presented in Figure \ref{fig_cfdqn} and Algorithm 2, respectively. In each episode of the agent-environment interactive process, a user $u$ is uniformly sampled from training set $\mathcal{U}_{train}$ as the environment, which will be used to interact with the agent for $T$ time steps, and to generate corresponding $T$ transitions $(s_t, a_t, r_{t+1}, s_{t+1})$ for $t=0,...,T-1$ based on $u$'s data. At each time step $t$, the agent observes reward $r_{t+1}$ and raw state $s_{t+1}^o$ from the environment after executing action $a_t$. Then, the CF-based state $s_{t+1}$ is computed by Algorithm 1, and the transition $(s_t,a_t,r_{t+1},s_{t+1})$ is added into an experience replay memory $\mathcal{M}$. When performing Q-learning updates, in stead of single transition, a minibatch of transitions is uniformly sampled from $\mathcal{M}$ to update the Q-network. Moreover, to ensure exploration at each time step $t$, the action $a_t$ is chosen by using a $\epsilon$-greedy strategy with regard to the predicted $Q$ values. The training process can continue for any number of episodes as long as the Q-network is not converged. After training, the learned CFRL agent can be used to make $T$-step interactive recommendations for any new user. The agent only needs to interact with the user step by step, observe states, and always take greedy actions with respect to the Q values outputted by the learned Q-network.

\begin{table}[t]
	\renewcommand\tabcolsep{1pt}
	\renewcommand\arraystretch{1.0}
	\centering
	\begin{tabular*}{\columnwidth}{rl} 
		\toprule
		\multicolumn{2}{l}{\textbf{Algorithm 2:} The CFRL Learning Algorithm} \\
		\hline
		\multicolumn{2}{l}{\textbf{Input:} training set $\mathcal{U}_{train}$, rating data $R$, the number of episodes $K$, the number of } \\
		\multicolumn{2}{l}{\qquad\quad time steps $T$, discount factor $\gamma$, $\epsilon$-greedy parameter $\epsilon$, replay memory $\mathcal{M}$} \\
		\multicolumn{2}{l}{\textbf{Output:} the learned Q-network $\hat{Q}$} \\
		1.& Initialize $\hat{Q}$ with random weights \\
		2.& \textbf{for} $\text{episode}=1,...,K$ \textbf{do} \\
		3.& \quad Uniformly pick a user $u\in\mathcal{U}_{train}$ as the environment \\
		4.& \quad Observe raw state $s_0^o$ \\
		5.& \quad Compute CF-based state $s_0$ by Algorithm 1 \\
		6.& \quad \textbf{for} $t=0,...,T-1$ \textbf{do} \\
		7.& \qquad Select action $a_t$ using $\epsilon$-greedy policy w.r.t. $\hat{Q}$ \\
		8.& \qquad Take $a_t$, observe reward $r_{t+1}$ and raw state $s_{t+1}^o$ \\
		9.& \qquad Compute CF-based state $s_{t+1}$ by Algorithm 1 \\
		10.& \qquad Store transition $(s_t, a_t, r_{t+1}, s_{t+1})$ in $\mathcal{M}$ \\
		11.& \qquad Sample a minibatch of $(s, a, r, s')$ from $\mathcal{M}$ \\
		12.& \qquad Update $\hat{Q}$'s weights $\textbf{w}$ according to Equation \ref{eq_qlearning_fa} \\
		13.& \quad \textbf{end for} \\
		14.& \textbf{end for} \\
		\hline
	\end{tabular*}
\end{table}

\textbf{Time Complexity Analysis}. We now analyze the time complexity of the CFRL learning algorithm. In the inner for loop in Algorithm 2, the time is mainly taken by computing the predicted Q values (line 7), computing CF-based state $s_{t+1}$ (line 9), and updating $\hat{Q}$'s weights (line 12). The costs of computing Q values and updating $\hat{Q}$'s weights are both $O(|\textbf{w}|)$, where $|\textbf{w}|$ denotes the number of $\hat{Q}$'s weights. The cost of computing CF-based state is $O(d)$ (since only one iteration is needed for the while loop in Algorithm 1), where $d$ is the dimensionality of latent feature vectors. Therefore, the time complexity of the CFRL learning algorithm is $O(KT(d+|\textbf{w}|))$, where $K$ is the number of episodes and $T$ is the number of time steps.

\section{Experiments}
\subsection{Experimental Settings}
\textbf{Datasets}.
We employ three benchmark explicit-feedback datasets from MovieLens\footnote{http://grouplens.org/datasets/movielens/}: ML100K, ML1M and ML10M. All the datasets contain integer ratings (from 1 to 5) of movies given by users, and each user has at least 20 observed ratings. The average value of ratings of the three datasets are 3.529, 3.581 and 3.512, respectively.

\textbf{Evaluation Protocol}.
To evaluate interactive recommendation algorithms, we follow an unbiased offline evaluation scheme suggested in previous work \cite{zhao2013interactive,li2010contextual} that the pre-collected ratings are treated as unbiased interactive feedback of users. We regard those users who have more than 100 ratings as candidates for testing purpose. For each dataset, we split the data by randomly choosing 10\% candidates as testing set $\mathcal{U}_{test}$, and the remaining as training set $\mathcal{U}_{train}=\mathcal{U}\setminus\mathcal{U}_{test}$. We repeat the above process 10 times independently and obtain 10 data splits. We conduct each experiment based on the 10 data splits and report the average results for evaluation. The evaluation metric we used is the average reward (rating) received over $T$ time steps. 

Moreover, how to regard the unknown (missing) ratings is non-trivial for evaluating recommendation algorithms. In the literature, there are two widely-used strategies. The unknown ratings are usually ignored in rating prediction task \cite{Koren2009}, and regarded as negative feedback in top-n recommendation task \cite{CremonesiKorenTurrin2010}. To comprehensively evaluate CFRL, we adopt both of the strategies and derive two different tasks for $T$-step interactive recommendation:
\begin{itemize}
	\item \textbf{Task I}. We ignore the unknown ratings. The possible actions (items) available for the agent are restricted in the set of rated items of the target user. 
	\item \textbf{Task II}. We regard the unknown ratings as negative feedback. The entire item set is available for the agent. The reward of recommending an item with unknown rating is defined as 0. 
\end{itemize}

Using movie recommendation as an example, Task I focuses on predicting how much the user will like a movie, while Task II aims at predicting both whether the user will watch a movie, and how much she will like it. In general, Task II is more difficult than Task I. For both tasks, we set $T=40$ for our evaluation.

\textbf{Baselines for Comparison}.
We compare CFRL against a wide variety of existing approaches, including active learning (AL), collaborative filtering (CF), multi-armed bandit (MAB), and reinforcement learning (RL) methods. We summarize the brief description of all baselines below.

\begin{itemize}
	\item \textbf{Random}. This method randomly picks an item from the available set at each time step.
	\item AL methods: \textbf{Popular} \cite{rashid2002getting} and \textbf{Impact} \cite{mello2010active}. The Popular method picks the most popular item. The Impact method picks the item which has highest impact on other items, where the impact is computed based on a bipartite graph of users and items. 
	\item CF method: \textbf{MF} \cite{Koren2009}. This method is a standard matrix factorization model with online updates by stochastic gradient descent. 
	\item MAB method: \textbf{LinUCB} \cite{li2010contextual}. The LinUCB method is a contextual MAB algorithm for news recommendation. We extend the original LinUCB for our interactive recommendation problem. We adopt the same training scheme of CFRL for LinUCB, and concatenate CF-based state and item vector as the context vector of LinUCB.
	\item RL method: \textbf{DQN} \cite{mnih2015human}. The DQN method uses the same Q-network and training scheme of CFRL. However, its Q-network takes the raw states as input, rather than the CF-based states used by CFRL.
\end{itemize}

\subsection{Experimental Results}
We now compare the overall performance of all methods, in terms of the average reward received over $T=40$ time steps. The mean and standard deviation of the results over 10 runs on datasets ML100K and ML1M are reported in Table \ref{table_results}. For the largest dataset ML10M, we only report the results over 1 run due to our limited computing resources. In each case in Table \ref{table_results}, the bold font indicates the best performing method, and the mark $*$ denotes the second-best performing one. The $p$-value is computed by conducting paired $t$-test for the two methods. The relative improvement of the best method over the second-best one is also shown in the last row. The proposed CFRL outperforms the baselines remarkably in all cases. In particular, for the more difficult Task II, the improvements of CFRL over the best performing baselines are 14.58\%, 19.90\% and 8.76\% on datasets ML100K, ML1M and ML10M, respectively.

\begin{table*}[t]
	\renewcommand\tabcolsep{2pt}
	\renewcommand\arraystretch{1.0}
	\centering
	\caption{Overall comparison in terms of the average reward over $T=40$ steps.}
	\begin{tabular}{llllllll}
		\toprule
		\multirow{2}{*}{\bf Methods} &\multicolumn{3}{c}{\bf Task I} &&\multicolumn{3}{c}{\bf Task II} \\
		\cline{2-4}\cline{6-8}
		&\bf ML100K &\bf ML1M &\bf 10M  &&\bf ML100K &\bf ML1M  &\bf 10M \\
		\hline
		Random  &3.513$\pm$0.066 &3.611$\pm$0.019 &3.561 &&0.454$\pm$0.031 &0.272$\pm$0.017 &0.016 \\
		Popular &3.837$\pm$0.061 &4.035$\pm$0.021 &3.798 &&2.404$\pm$0.095 &2.411$\pm$0.036*&2.126 \\
		Impact  &3.870$\pm$0.054 &4.015$\pm$0.023 &3.851 &&2.634$\pm$0.089*&2.397$\pm$0.043 &2.269* \\
		MF    &4.059$\pm$0.050*&4.207$\pm$0.013 &4.069 &&1.979$\pm$0.068 &1.730$\pm$0.055 &1.735 \\
		LinUCB  &4.053$\pm$0.059 &4.214$\pm$0.015*&4.078*&&2.544$\pm$0.083 &2.360$\pm$0.033 &2.190 \\
		DQN     &3.971$\pm$0.071 &4.058$\pm$0.007 &3.945 &&1.786$\pm$0.082 &1.352$\pm$0.088 &0.747 \\
		CFRL &\bf4.105$\pm$0.047 &\bf4.274$\pm$0.016 &\bf4.156 &&\bf3.018$\pm$0.085 &\bf2.891$\pm$0.058 &\bf2.468 \\
		\hline
		$p$-value &0.002  &2e-5   &-        &&4e-5    &5e-6     &- \\
		Improve   &1.13\% &1.41\% &1.91\%   &&14.58\% &19.90\%  &8.76\%  \\
		\hline
	\end{tabular}
	\label{table_results}
\end{table*}

\section{Conclusions}
In this paper, we study an interactive recommendation problem for explicit-feedback recommender systems. We develop a novel and effective approach, named CFRL, which seamlessly integrates the ideas of both CF and RL. We first model the interactive recommendation problem as a standard RL task, with a novel CF-based MDP which makes the collaborative recommendations available. We then develop an effective Q-network learning method to learn the agent's optimal policy based on the CF-based MDP. The capability of CFRL is demonstrated by the comprehensive experimental results and analysis.

%\section*{Acknowledgments}

%The bibliography
\bibliographystyle{splncs04}
\bibliography{leiyu}

\begin{thebibliography}{10}
\providecommand{\url}[1]{\texttt{#1}}
\providecommand{\urlprefix}{URL }
\providecommand{\doi}[1]{https://doi.org/#1}

\bibitem{CremonesiKorenTurrin2010}
Cremonesi, P., Koren, Y., Turrin, R.: Performance of recommender algorithms on
  top-n recommendation tasks. In: Proceedings of the fourth ACM conference on
  Recommender systems. pp. 39--46. ACM (2010)

\bibitem{he2016fast}
He, X., Zhang, H., Kan, M.Y., Chua, T.S.: Fast matrix factorization for online
  recommendation with implicit feedback. In: Proceedings of the 39th
  International ACM SIGIR conference on Research and Development in Information
  Retrieval. pp. 549--558. ACM (2016)

\bibitem{hu2017playlist}
Hu, B., Shi, C., Liu, J.: Playlist recommendation based on reinforcement
  learning. In: International Conference on Intelligence Science. pp. 172--182.
  Springer (2017)

\bibitem{Koren2009}
Koren, Y., Bell, R., Volinsky, C.: Matrix factorization techniques for
  recommender systems. Computer (8),  30--37 (2009)

\bibitem{li2010contextual}
Li, L., Chu, W., Langford, J., Schapire, R.E.: A contextual-bandit approach to
  personalized news article recommendation. In: Proceedings of the 19th
  international conference on World wide web. pp. 661--670. ACM (2010)

\bibitem{mello2010active}
Mello, C.E., Aufaure, M.A., Zimbrao, G.: Active learning driven by rating
  impact analysis. In: Proceedings of the fourth ACM conference on Recommender
  systems. pp. 341--344. ACM (2010)

\bibitem{mnih2015human}
Mnih, V., Kavukcuoglu, K., Silver, D., Rusu, A.A., Veness, J., Bellemare, M.G.,
  Graves, A., Riedmiller, M., Fidjeland, A.K., Ostrovski, G., et~al.:
  Human-level control through deep reinforcement learning. Nature
  \textbf{518}(7540),  529--533 (2015)

\bibitem{rashid2002getting}
Rashid, A.M., Albert, I., Cosley, D., Lam, S.K., McNee, S.M., Konstan, J.A.,
  Riedl, J.: Getting to know you: learning new user preferences in recommender
  systems. In: Proceedings of the 7th international conference on Intelligent
  user interfaces. pp. 127--134. ACM (2002)

\bibitem{ricci2011introduction}
Ricci, F., Rokach, L., Shapira, B.: Introduction to recommender systems
  handbook. In: Recommender systems handbook, pp. 1--35. Springer (2011)

\bibitem{shani2005mdp}
Shani, G., Heckerman, D., Brafman, R.I.: An mdp-based recommender system.
  Journal of Machine Learning Research  \textbf{6}(Sep),  1265--1295 (2005)

\bibitem{silver2016mastering}
Silver, D., Huang, A., Maddison, C.J., Guez, A., Sifre, L., Van Den~Driessche,
  G., Schrittwieser, J., Antonoglou, I., Panneershelvam, V., Lanctot, M.,
  et~al.: Mastering the game of go with deep neural networks and tree search.
  Nature  \textbf{529}(7587),  484--489 (2016)

\bibitem{sutton1998reinforcement}
Sutton, R.S., Barto, A.G.: Reinforcement learning: An introduction, vol.~1. MIT
  press Cambridge (1998)

\bibitem{zhao2018deep}
Zhao, X., Xia, L., Zhang, L., Ding, Z., Yin, D., Tang, J.: Deep reinforcement
  learning for page-wise recommendations. arXiv preprint arXiv:1805.02343
  (2018)

\bibitem{zhao2018recommendations}
Zhao, X., Zhang, L., Ding, Z., Xia, L., Tang, J., Yin, D.: Recommendations with
  negative feedback via pairwise deep reinforcement learning. arXiv preprint
  arXiv:1802.06501  (2018)

\bibitem{zhao2013interactive}
Zhao, X., Zhang, W., Wang, J.: Interactive collaborative filtering. In:
  Proceedings of the 22nd ACM international conference on Conference on
  information \& knowledge management. pp. 1411--1420. ACM (2013)

\bibitem{zheng2018drn}
Zheng, G., Zhang, F., Zheng, Z., Xiang, Y., Yuan, N.J., Xie, X., Li, Z.: Drn: A
  deep reinforcement learning framework for news recommendation. In: TheWebConf
  2018. ACM (2018)

\end{thebibliography}

\end{document}